%% file: main.tex
\documentclass[letterpaper]{article} 
\usepackage[]{aaai25}  
\usepackage{times}  
\usepackage{helvet}  
\usepackage{courier}  
\usepackage[hyphens]{url}  
\usepackage{graphicx} 
\usepackage{natbib}  
\usepackage{caption} 
\usepackage{algorithm}
\usepackage{algpseudocode}      

\usepackage{amsmath}            
\usepackage{amssymb}            
\usepackage{amsfonts}           
\usepackage{multirow}           
\usepackage{booktabs}           
\usepackage{makecell}           
\usepackage{tabularx}           
\usepackage{arydshln}           
\usepackage{color}              
\usepackage{CJKutf8}            

\usepackage{newfloat}
\usepackage{listings}
\DeclareCaptionStyle{ruled}{labelfont=normalfont,labelsep=colon,strut=off} 
\floatstyle{ruled}
\newfloat{listing}{tb}{lst}{}
\floatname{listing}{Listing}
\title{EXCGEC: A Benchmark for Edit-Wise Explainable Chinese\\Grammatical Error Correction}

\author{
    Jingheng Ye\textsuperscript{\rm 1}\equalcontrib,
    Shang Qin\textsuperscript{\rm 1}\equalcontrib,
    Yinghui Li\textsuperscript{\rm 1},
    Xuxin Cheng\textsuperscript{\rm 2},
    Libo Qin\textsuperscript{\rm 3},
    Hai-Tao Zheng\textsuperscript{\rm 1}\thanks{Corresponding Authors.},\\
    Ying Shen\textsuperscript{\rm 4},
    Peng Xing\textsuperscript{\rm 1},
    Zishan Xu\textsuperscript{\rm 1},
    Guo Cheng\textsuperscript{\rm 1},
    Wenhao Jiang\textsuperscript{\rm 5 $\dagger$} \\
}
\affiliations{
    \textsuperscript{\rm 1}Tsinghua University,\\
    \textsuperscript{\rm 2}Peking University,\\
    \textsuperscript{\rm 3}Central South University,\\
    \textsuperscript{\rm 4}Sun Yat-Sen University,\\
    \textsuperscript{\rm 5}Guangdong Laboratory of Artificial Intelligence and Digital Economy (SZ)\\
    \{yejh22, qin-s23, liyinghu20\}@mails.tsinghua.edu.cn
}

\begin{document}
\begin{CJK*}{UTF8}{gbsn}

\maketitle

\definecolor{Evidence}{RGB}{255,0,0}
\definecolor{Linguistic}{RGB}{84,130,53}
\definecolor{Cause}{RGB}{197,90,17}
\definecolor{Revision}{RGB}{31,78,121}

\newcommand{\Task}{EXGEC}
\newcommand{\Model}{EXGEC}
\newcommand{\Benchmark}{\textsc{EXCGEC}}
\newcommand{\Decode}{COTE}

\input{chapters/abstract}
\input{chapters/introduction}
\input{chapters/related_work}

\input{chapters/method}
\input{chapters/experiments}
\input{chapters/conclusion}

\bibliography{aaai25}

\input{chapters/appendix}

\end{CJK*}
\end{document}

%% file: chapters/abstract.tex
\begin{abstract}

Existing studies explore the explainability of Grammatical Error Correction (GEC) in a limited scenario, where they ignore the interaction between corrections and explanations and have not established a corresponding comprehensive benchmark. To bridge the gap, this paper first introduces the task of EXplainable GEC (\textbf{\Task}), which focuses on the integral role of correction and explanation tasks. To facilitate the task, we propose \textbf{EXCGEC}, a tailored benchmark for Chinese EXGEC consisting of 8,216 explanation-augmented samples featuring the design of hybrid edit-wise explanations. We then benchmark several series of LLMs in multi-task learning settings, including post-explaining and pre-explaining. To promote the development of the task, we also build a comprehensive evaluation suite by leveraging existing automatic metrics and conducting human evaluation experiments to demonstrate the human consistency of the automatic metrics for free-text explanations. Our experiments reveal the effectiveness of evaluating free-text explanations using traditional metrics like METEOR and ROUGE, and the inferior performance of multi-task models compared to the pipeline solution, indicating its challenges to establish positive effects in learning both tasks.
\end{abstract}

\begin{links}
\link{Code \& Data}{https://github.com/THUKElab/EXCGEC}
\end{links}

%% file: chapters/introduction.tex
\section{Introduction}\label{sec:introduction}


Despite the notable advancements in Grammatical Error Correction (GEC)~\cite{bryant2023grammatical,DBLP:conf/emnlp/YeLL023, LI2025126039}, there still exists a lack of profound examination into the explainability of GEC~\cite{dwivedi2023explainable}, which is critical in educational scenarios for L2 (second language)-speakers~\cite{wang2021yaclc}. These mainstream users, who often face challenges in creating grammatically accurate and fluent texts, may be confused or even misguided if provided with limited access to only corrective texts~\cite{ye2025position}. Therefore, augmenting the explainability of GEC is unquestionably beneficial for the progression of GEC as well as related fields, such as essay scoring~\cite{stahl2024exploring}, intelligent tutoring systems~\cite{montenegro2023impact}.

\input{figures/intro}

As illustrated in Figure~\ref{fig:intro}, existing tasks like GEC~\cite{ye2022focus} and Grammatical Error Explanation (GEE) typically address either correction or explanation, ignoring the interaction between the two. To bridge the gap, we introduce the task of \textbf{EX}plainable \textbf{G}rammatical \textbf{E}rror \textbf{C}orrection (\textbf{\Task}). By integrating these two tasks,~\Task~enables systems to elucidate the linguistic knowledge and reasoning mechanism underlying predicted corrections, thus achieving the best of both worlds. Additionally,~\Task~can function as a test bed for determining the explainable abilities of large language models (LLMs) and identifying any unintended biases and risks in educational scenarios.

To facilitate~\Task, we present \textbf{EXCGEC}, a tailored benchmark for Chinese~\Task, featuring the design of hybrid edit-wise explanations. Each explanation, based on a particular edit, consists of three elements:
1) \textit{Error types}, which allow learners to absorb syntax and semantic knowledge in an inductive way~\cite{fei-etal-2023-enhancing}. We establish a hierarchical and pragmatic two-tier taxonomy for Chinese grammatical errors.
2) \textit{Error severity levels} ranging from 1 $\sim$ 5 points, which are beneficial to prioritize core corrections.
3) \textit{Error descriptions}, presented as the form of natural language explanation~\cite{camburu2018snli,he2023using}, provide evidence words, relevant linguistic knowledge or syntax rules, error causes, and revision advice for edits.
The edit-wise design provides more detailed and faithful guidance for learners, allowing them to comprehend each grammatical error committed. This is unlikely achievable for other designs such as example-based~\cite{kaneko-etal-2022-interpretability} or sentence-level explanations~\cite{nagata2021shared}.

Stimulated by the recent success of synthetic data generation~\cite{shum-etal-2023-automatic,whitehouse-etal-2023-llm}, we employ a semi-automatic dataset construction solution to enhance efficiency, while minimising annotation costs. Initially, we synthesize the evaluation part of EXCGEC by prompting GPT-4~\cite{liu2024best}. Then we hire experienced annotators to filter out invalid data and concurrently provide a detailed analysis of the invalid data, ensuring the quality of our dataset~\cite{ding2024data}. We finally obtain 8,216 clean explanation-augmented samples for benchmarking.
Additionally, We utilize existing automatic metrics to evaluate the performance. Particularly for error descriptions, we conduct a human evaluation experiment to ascertain the correlation between the metrics and human judgements, thus demonstrating their effectiveness.



Based on the benchmark, we develop EXGEC multi-task baseline models that can perform both the correction and explanation tasks in either post-explaining (correct-then-explain) or pre-explaining (explain-then-correct) sequences. Particularly, we design \textbf{Co}rrect-\textbf{T}hen-\textbf{E}xplain (\textbf{COTE}) decoding algorithm for post-explaining models. Benchmarking various series of open-source LLMs has yielded several intriguing findings. For example, post-explaining models display higher performance than pre-explaining models. However, both of them under-perform the pipeline solution.
Moreover, COTE significantly enhances performance by alleviating the alignment workload for the LLMs. Our contributions in this paper are listed as follows:

\begin{itemize}
    \item We introduce the EXGEC task and establish a corresponding benchmark consisting of a Chinese EXGEC dataset and a comprehensive set of metrics, contributing to the stable development of the field of~\Task.

    \item We develop EXGEC baseline models and investigate the abilities of various LLMs using our proposed benchmark.

    \item We conduct detailed analyses on our proposed dataset and baselines to gain further insights. Human evaluation experiments are also conducted to confirm the effectiveness of automatic metrics for error descriptions.
\end{itemize}

%% file: figures/intro.tex
\begin{figure}[tb!]
\centering
\includegraphics[scale=0.25]{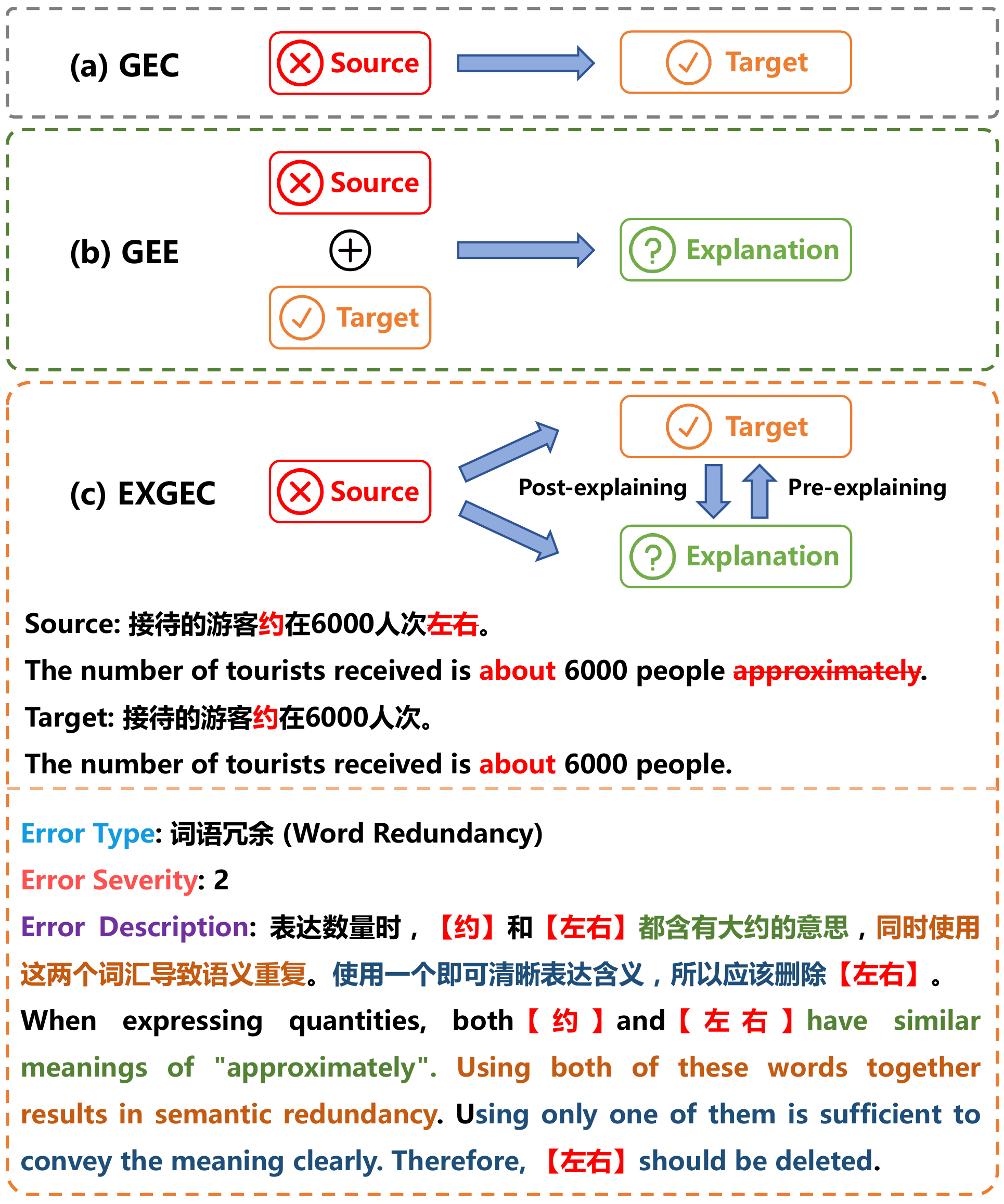}
\caption{
Task definitions of GEC, GEE, and EXGEC. We highlight \textcolor{Evidence}{【evidence words】}, \textcolor{Evidence}{\{correction\}}, \textcolor{Linguistic}{linguistic knowledge}, \textcolor{Cause}{error causes}, and \textcolor{Revision}{revision advice} parts.}
\label{fig:intro}
\end{figure}

%% file: chapters/related_work.tex
\section{Related Work}\label{sec:related_work}


\paragraph{Explainable GEC.}
Exploration of explainable GEC has witnessed a paradigm shift from fine-tuning to prompting~\cite{zhao2024explainability}. EXPECT~\cite{fei-etal-2023-enhancing} is an explainable GEC dataset annotated with evidence words and error types based on the standard GEC benchmark~\cite{bryant-etal-2019-bea}. However, EXPECT falls short of flexibility due to the lack of natural language explanations. To fill the gap,~\citet{song2023gee} propose the task of grammatical error explanation. They observe that GPT-4 suffers from identifying and explaining errors with limited access to only parallel source-target pairs. To address this issue, they fine-tune an extra LLM as an edit extractor trained on synthesized data. On the other hand, a similar task called feedback comment generation, focuses on sentence-level explanations. However, it suffers from expensive costs associated with data annotation~\cite{nagata2020creating}. Furthermore, it is explored with limited access to only a subset of English grammatical error types due to the complexity of the task~\cite{nagata2019toward}. In conclusion, all these studies do not establish a comprehensive benchmark integrating both the tasks of GEC and GEE, and thus lack in-depth exploration in multi-task learning the both tasks. However, our work is the first to propose a systematic framework for EXCGEC.

\paragraph{Chinese GEC.} The research on CGEC~\cite{DBLP:conf/emnlp/YeLL023,ye-etal-2023-system} has also come a long way recently, along with a series of CGEC datasets~\cite{zhao2018overview}. Similar to those in English, Chinese grammatical errors can also be categorized into different error types. CLG~\cite{ma-etal-2022-linguistic} divides Chinese grammatical errors into 6 categories: Structural Confusion, Improper Logicality, Missing Component, Redundant Component, Improper Collocation, and Improper Word Order. However, the taxonomy of CLG is targeted toward grammatical errors made by native speakers and thereby can not cover those made by L2 speakers. To fill the gap, we design a two-tier hierarchical taxonomy, which is capable of covering most grammatical errors.





%% file: chapters/method.tex
\section{Task Definition}\label{sec:task}
\subsection{Grammatical Error Correction}\label{subsec:gec}
GEC~\cite{schneider-mccoy-1998-recognizing-syntactic} has been studied for decades, witnessing the shift from rule-based methods to LLM-based methods. Formally, given an ungrammatical text (source text) $X=\{x_1,x_2,\cdots,x_T\}$, a GEC model is required to correct $X$ into a grammatically correct counterpart (target text) $Y=\{y_1,y_2,\cdots,y_{T'}\}$ without changing the original semantic as far as possible. Typically, GEC is usually treated as a sequence-to-sequence (Seq2Seq) task, the training objective of which is formulated as follows:

\begin{equation}\begin{aligned}
    \mathcal{L}_{\operatorname{GEC}} = -\sum_{t=1}^{T'} \log P(y_t \mid Y_{<t}, X).
\end{aligned}\end{equation}

\subsection{Grammatical Error Explanation}\label{subsec:gee}
GEE~\cite{song2023gee} has received much attention recently and has been explored in several methodologies, including sentence-level explanation and edit-wise explanation. Since sentence-level explanations suffer from over-generalization and confusion especially when a sentence contains multiple grammatical errors, this work focuses solely on edit-wise explanations. Given a source text $X$ and its target counterpart $Y$, the GEE model needs to explain each grammatical error $e_i$ in $X$. Specifically, GEE is typically solved in a two-step pipeline consisting of edit extraction and edit-wise explanation.
1) \textbf{Edit extraction} produces an edit set $E=\{e_1,e_2,\cdots,e_n\}$ that represent grammatical errors in $X$ and also clarify the transformation from ungrammatical segments of $X$ to target segments of $Y$. Typically, an edit contains four key elements: source position $sp$, source content $sc$, target position $tp$, and target content $tc$. The process of edit extraction can be easily accomplished using alignment-based evaluation toolkits like ERRANT~\cite{bryant-etal-2017-automatic} and CLEME~\cite{ye-etal-2023-cleme,ye2024cleme2}.
2) \textbf{Edit-wise explanation} generates a set of explanations $E'=\{e'_1,e'_2,\cdots,e'_n\}$, with each explanation $e'_i$ corresponding to $e_i$, given $X$ and $Y$. Although the design of explanation varies across related work~\cite{song2023gee,zhao2024explainability}, the typical training objective of GEE models is presented as follows:

\begin{equation}\begin{aligned}
    E = f(X, Y),
\end{aligned}\end{equation}
\begin{equation}\begin{aligned}
    \mathcal{L}_{\operatorname{GEE}} = -\sum_{i=1}^{n} \log P(e'_i \mid X, Y, e_i),
\end{aligned}\end{equation}
where $f: (X, Y) \to E = \{(sp_i, sc_i, tp_i, tc_i)\}^n_{i=1}$ is the edit extraction function used to extract edits of $X$ and $Y$, and $n$ is the number of edits.

Existing studies~\cite{song2023gee,fei-etal-2023-enhancing} focus on developing GEE models that can generate explanations. However, an extra GEC model is compulsory for GEE models to work, thus resulting in an issue of low efficiency.




\input{figures/overview}

\subsection{Explainable Grammatical Error Correction}\label{subsec:exgec}
To get rid of the drawbacks brought by the nature of GEE, we propose the EXGEC task which aims to perform both correction and explanation tasks simultaneously. The motivation for combining these two tasks majorly falls on two aspects. First, a branch of existing studies~\cite{Wiegreffe2021teach,hartmann-sonntag-2022-survey,li2022unifying,li2024explanations} have demonstrated training with access to human explanations can improve model performance. It is also intuitive that either of the GEC and GEE tasks can mutually benefit from each other when training in a multi-task manner. Second, it is more time-saving and cost-efficient to deploy a single EXGEC model rather than two detached models in foreign language education platforms.

In this task, the only input element is an ungrammatical source text $X$, and the EXGEC model learns to output both the grammatical target text $Y$ and explanations $E'$. Similar to GEE, EXGEC follows the edit-wise style of explanation, and it is categorized into two different settings by the order of correction and explanation tasks, with the basic scheme of multi-task learning.

\paragraph{Post-explaining.} Models are trained first to generate target texts~\cite{camburu2018snli}, which allows the explanations to be explicitly conditioned on the target texts, thus ensuring high faithfulness of explanations towards the target texts. The training objective is as follows:

\begin{equation}\begin{small}\begin{aligned}
    \mathcal{L}_{\operatorname{post}} =& -\sum_{t=1}^{T'} \log P(y_t \mid Y_{<t}, X) -\sum_{i=1}^{n} \log P(e'_i \mid X, Y, e_i).
\end{aligned}\label{eq:post}\end{small}\end{equation}

The inference of post-explaining models is as follows:

\begin{equation}\begin{aligned}
    \hat{Y} = \operatorname{EXGEC}_{\operatorname{post}}(X),
\end{aligned}\end{equation}
\begin{equation}\begin{aligned}
    \hat{E}' = \operatorname{EXGEC}_{\operatorname{post}}(X, Y, f(X, \hat{Y})).
\end{aligned}\end{equation}

With the target texts generated ahead, post-explaining models can output explanations conditioned on the specific edits extracted by an aligning process, thus improving the accuracy and faithfulness of explanations.

\paragraph{Pre-explaining.} This type of model is trained in converse order, whose mechanism is similar to the Chain of Thought (CoT) technique. Pre-explaining models are supposed to make full use of synthesized explanations to generate elaborated target texts. With minimal modification from Equation~\eqref{eq:post}, the training objective of pre-explaining models is as follows:

\begin{equation}\begin{small}\begin{aligned}
    \mathcal{L}_{\operatorname{pre}} =& -\sum_{i=1}^{n} \log P(e'_i \mid X) -\sum_{t=1}^{T'} \log P(y_t \mid Y_{<t}, X, E').
\end{aligned}\end{small}\end{equation}

Notably, pre-explaining models may struggle to generate well-formed edit-wise explanations due to the inaccessibility to the edit extraction function $f$, which necessitates both the source and the target texts. Similarly, the inference of pre-explaining models is presented as follows:

\begin{equation}\begin{aligned}
    \hat{E}' = \operatorname{EXGEC}_{\operatorname{pre}}(X),
\end{aligned}\end{equation}
\begin{equation}\begin{aligned}
    \hat{Y} = \operatorname{EXGEC}_{\operatorname{pre}}(X, E').
\end{aligned}\end{equation}

\section{\Benchmark~Benchmark}\label{sec:dataset}
To facilitate the development of~\Task~task, we construct~\Benchmark, the first benchmark for explainable Chinese GEC particularly. As illustrated in Figure~\ref{fig:overview}, we begin with the process of data curation, which consists of Explanation Design, Explanation Synthesizing, Explanation Refinement, and Analysis. Then we gain an in-depth understanding of GPT-4~\cite{achiam2023gpt} by further analyzing the generated explanations, where we summarize common failure modes in invalid instances. Finally, we explain the evaluation for both the correction and the explanation tasks.

\input{tables/error_types}

\subsection{Explanation Design}\label{subsec:exp_design}
In the pursuit of comprehensiveness and plausibility, we adopt a hybrid strategy for edit-wise explanations, where each edit is explained through three aspects, including error type labels, error severity levels, and free-text error descriptions.
1) \textbf{Error type labels} allow language learners to comprehend and inductively infer syntax and grammar rules. In particular, we employ a two-tier hierarchical taxonomy including 5 major types and 16 minor types shown in Table~\ref{tab:error_types}, inspired by authoritative linguistic books~\cite{borong2011modern,jingmin2016general}. Detailed descriptions of various error types are included in the supplementary materials. If an edit covers multiple error types, we select the one with the highest granule.
2) \textbf{Error severity levels}, ranging from 1 to 5 points, indicate the significance of a specific grammatical error.
3) \textbf{Error descriptions} are the most crucial and flexible element. These provide keywords, pertinent linguistic knowledge, causes of errors, and revision guidance in a free-text format. We stipulate well-defined error descriptions should meet three nonoverlapping principles: fluency, reasonability (making sense to humans), and faithfulness (targeted to a specific edit). To ensure reasonability and faithfulness, the error description must mostly conform to the syllogism form of deductive reasoning: \textit{[major premise: semantic rules and related knowledge]}, \textit{[minor premise: the reason for the error in the text]}, and \textit{[explain how to correct it]}. Further, any evidence from the source $X$ must be enclosed within special markers 【~】. Similarly, correction content that occurs in the target sentence $Y$ must be enclosed within \{~\}, as indicated in Figure~\ref{fig:intro}.


\subsection{Explanation Synthesizing}
\label{subsec:exp_creation}
Annotating high-quality explanations on a large scale poses a huge challenge to our benchmark construction. Hence, we leverage GPT-4 to synthesize edit-wise explanations efficiently. To achieve this, we first select 10,000 parallel samples across 6 existing benchmarks or datasets of Chinese GEC, including FCGEC~\cite{xu-etal-2022-fcgec}, YACLC~\cite{wang2021yaclc}, MuCGEC~\cite{zhang2022mucgec}, NaCGEC~\cite{ma-etal-2022-linguistic}, NLPCC~\cite{zhao2018overview} and HSK~\cite{zhang2009features}. The details are listed in Table~\ref{tab:statistics}. We pick out only the samples with \textit{changed} reference sentences to maximize training efficiency~\cite{zhang2022mucgec}. We select the reference sentence with the most edits as the target sentence if a sample is annotated with multiple reference sentences. Then, we prompt GPT-4 to generate edit-wise explanations following in-context learning. To ensure the faithfulness of the synthesized explanation, we first extract edits using the toolkit CLEME~\cite{ye-etal-2023-cleme}. Inspired by~\citet{li2022unifying}, we then employ the Rationalization Prompting (RP) strategy, where we concatenate task definition, demonstrations, and a parallel sample $(X, Y)$ with extracted edits $E=\{e_1,e_2,\cdots,e_n\}$ as the prompt. For each error type, we provide the definition, a suggested template of error description, and a demonstration. The prompt is listed in the supplementary materials.

\input{tables/statistics}



\subsection{Explanation Refinement and Analysis}
\label{subsec:data_analysis}
Benefiting from the extensive knowledge acquired during the large-scale pre-training process, GPT-4 can generate fluent, reasonable, and plausible explanations in most cases, meeting the requirements with specified instructions. However, GPT-4 is not guaranteed to produce all high-quality explanations due to hallucination, and the patterns of those invalid explanations are referred to as failure modes. Therefore, we hired 12 native speakers, all of whom are Chinese post-graduated students specializing in Chinese linguistics, to screen out invalid explanations. Before formal annotation, we compile the annotation guidelines and all the annotators receive intensive training. Two authors of the paper, who are also in charge of compiling the annotation guidelines, have made sure that their annotation accuracies are over 90\% on testing samples. We make sure that each formal sample is checked by at least two annotators. We finally obtained 8,216 clean samples out of 10,000 samples. We further investigate the failure modes of these invalid explanations, which are provided in the supplementary materials.


\subsection{Automatic Metrics}\label{subsec:metrics}
To promote the efficient development of EXGEC systems, we introduce a comprehensive suite of automatic metrics for both correction and explanation parts. Additionally, we conduct a human evaluation experiment in Section Analysis to demonstrate the alignment of the metrics used for assessing error descriptions with human judgments.

\paragraph{Correction.} We employ CLEME~\cite{ye-etal-2023-cleme} and ChERRANT~\cite{zhang2022mucgec} to evaluate the correction performance. Both are edit-based metrics that output P/R/F$_{0.5}$ scores, which have been proven reliable metrics for GEC on CoNLL-2014~\cite{ye-etal-2023-cleme}.


\paragraph{Explanation.} Since an edit-wise explanation consists of three critical elements, we define respectively automatic metrics for them.
1) Accuracy and Macro-F1 scores are computed for error type clarification, following the conventional evaluation protocol of text clarification~\cite{li2020survey}.
2) We report the mean absolute error (MAE) to show the deviation of hypothesis error severity levels towards ground truth ones.
3) We employ various metrics for evaluating the free-text explanation descriptions considering both the reproductivity and efficiency, including BLEU~\cite{papineni-etal-2002-bleu,clinciu-etal-2021-study}, METEOR~\cite{banerjee-lavie-2005-meteor}, ROUGE~\cite{lin-2004-rouge}.

\section{Method}\label{sec:method}
\paragraph{Training.} To streamline the training process covering all the tasks mentioned previously, we treat all of them as a unified Seq2Seq task. To achieve this, we linearize the data in the format of JSON~\cite{gao2023jsontuning}. This structured approach simplifies the process of output parsing involving three elements of edit-wise explanations, and provides a consistent and controllable view to distinguish tasks, enabling the model to understand essential task elements and their relations. With this uniform format stipulation, we can train all models using the same smooth cross-entropy loss, regardless of the specific task.

\begin{algorithm}[tbp!]
    \caption{\Decode~Decoding Algorithm}
    \label{alg:ransac}\small
    \renewcommand{\algorithmicrequire}{\textbf{Input:}}
    \renewcommand{\algorithmicensure}{\textbf{Output:}}
    \begin{algorithmic}[1]
        \Require
            Source text $X$, a post-explaining model $\mathcal{M}$, and the edit extraction function $f$.
        \Ensure
            Target text $\hat{Y}$, and explanations $\hat{E}'$.

        \State $\hat{Y} \gets \operatorname{BeamSearch}(\mathcal{M}(\operatorname{Json}(X)))$
        \State $\hat{E}' \gets \emptyset$
        \If{$\hat{Y} = X$}
        \State \Return $\hat{Y}, \hat{E}'$
        \EndIf
        \State $E \gets f(X, \hat{Y})$
        \State $\hat{E}' \gets \operatorname{Top-P}(\mathcal{M}(\operatorname{Json}(X, Y, E)))$
        \State \Return $\hat{Y}, \hat{E}'$
    \end{algorithmic}\label{alg}
\end{algorithm}

\paragraph{Inference.}\label{subsec:inference}
For post-explaining EXGEC models, we design a specific \textbf{Co}rrect-\textbf{T}hen-\textbf{E}xplain decoding algorithm called~\textbf{\Decode}, which is presented in Algorithm~\ref{alg}. First, we employ the greedy beam search decoding strategy for the correction part, which is beneficial to relieve the over-correction problem that is common in LLMs. Then, we apply CLEME to extract edits. Notably, we merge adjacent edits with a distance of less than 2 characters to avoid fragmented edits. Finally, we leverage the Top-p decoding strategy for generating explanations, encouraging diversified natural language explanations. It is worth noting that COTE is not accessible to pre-explaining models since the edit extraction tool necessitates both a source text and a target text.

%% file: figures/overview.tex
\begin{figure*}[tbp!]
\centering
\includegraphics[scale=0.26]{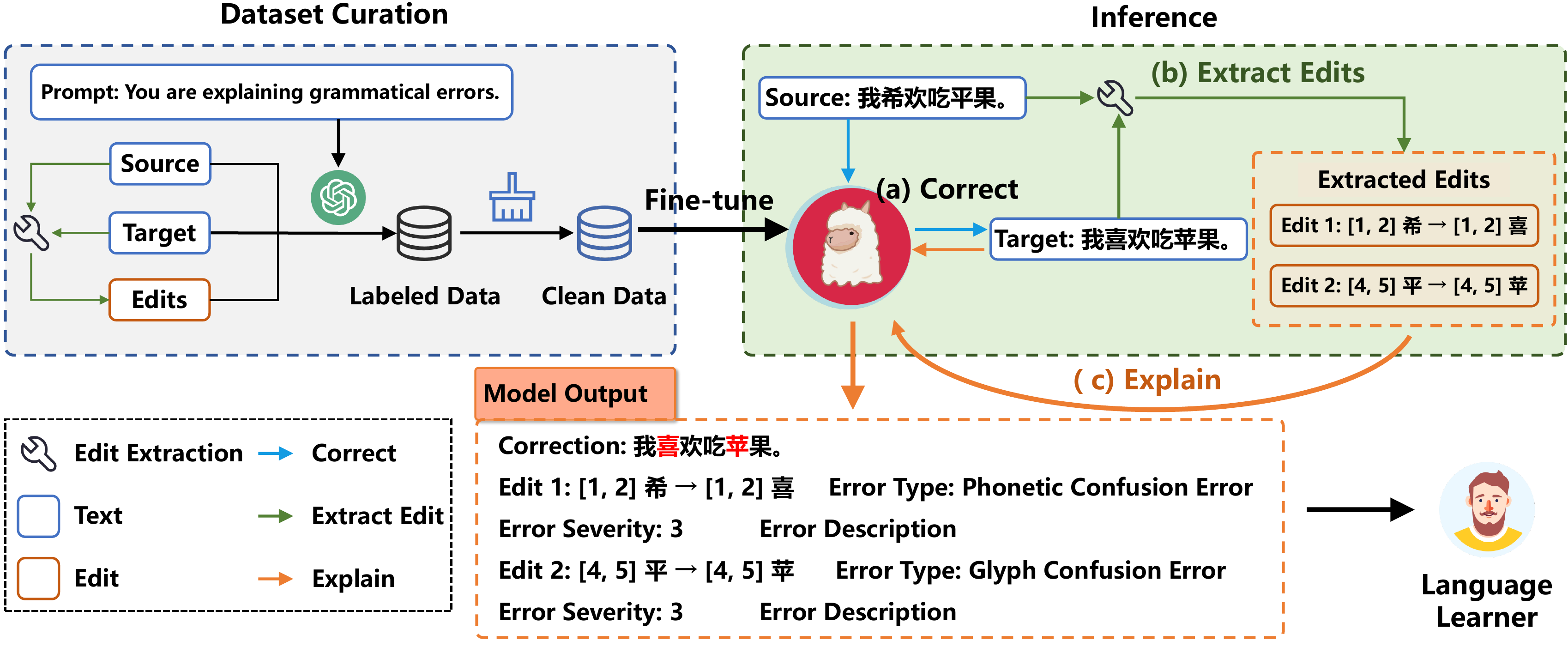}
\caption{Overview of the benchmark and the model. We show the inference process of a post-explaining model in particular.}
\label{fig:overview}
\end{figure*}

%% file: tables/error_types.tex
\begin{table}[htb!]
\renewcommand{\arraystretch}{0.9}
\renewcommand{\tabcolsep}{5pt}
\centering
\scriptsize
\begin{tabular}{ll}
    \toprule
    \textbf{Major Type}  &  \textbf{Minor Type}  \\
    \midrule
    
    \multirow{3}{*}{\textbf{Punctuation-level Error}}
    &  标点冗余~(Punctuation Redundancy)    \\
    &  标点丢失~(Punctuation Missing)       \\
    &  标点误用~(Punctuation Misuse)        \\

    \midrule

    \multirow{5}{*}{\textbf{Spelling-level Error}}
    &  字音混淆错误~(Phonetic Confusion Error)    \\
    &  字形混淆错误~(Glyph Confusion Error)       \\
    &  词内部字符异位错误                          \\
    &  (Internal Character Misplacement Error)   \\
    &  命名实体拼写错误~(Named Entity Misspelling) \\

    \midrule

    \multirow{3}{*}{\textbf{Word-level Error}}
    &  词语冗余~(Word Redundancy)    \\
    &  词语丢失~(Word Missing)       \\
    &  词语误用~(Word Misuse)        \\

    \midrule

    \multirow{3}{*}{\textbf{Sentence-level Error}}
    &  词序不当~(Improper Word Order)   \\
    &  逻辑不通~(Illogicality)          \\
    &  句式杂糅~(Run-on Sentence)       \\

    \midrule

    \multirow{3}{*}{\textbf{Other Special Error}}
    &  照应错误~(Inconsistency Error)   \\
    &  歧义错误~(Ambiguity Error)       \\
    &  语气不协调~(Inconsistent Tone)    \\

    \midrule
    \multicolumn{2}{c}{\textbf{Other}}  \\
    \bottomrule
\end{tabular}
\caption{Hierarchical taxonomy of grammatical error types.}
\label{tab:error_types}
\end{table}

%% file: tables/statistics.tex
\begin{table}[tbp!]
\renewcommand{\arraystretch}{0.9}
\renewcommand{\tabcolsep}{5pt}
\centering
\scriptsize
\begin{tabular}{lccc}
    \toprule
    
    \textbf{Dataset}  &  \textbf{Sentences}  &  \textbf{Edits/Sent.}  &  \textbf{Chars/Sent.}\\

    \midrule
    
    \textbf{FCGEC}  &  41,340  &  1.0  &  53.1  \\

    \textbf{YACLC-minimal-dev}  &  1,839  &  2.9  &  25.9  \\

    \textbf{MuCGEC-dev}  &  1,137  &  3.2  &  38.5  \\

    \textbf{NaCGEC-dev}  &  500  &  1.1  &  56.2  \\

    \textbf{NLPCC-test}  &  2,000  &  2.0  &  29.7  \\

    \textbf{HSK}  &  156,870  &  1.4  &  27.2  \\

    \midrule

    \textbf{EXCGEC} (FCGEC)  &  2,308  &  1.1  &  55.1  \\

    \textbf{EXCGEC} (YACLC)  &  1,235  &  3.5  &  24.3  \\

    \textbf{EXCGEC} (MuCGEC-dev)  &  789  &  3.3  &  40.4  \\

    \textbf{EXCGEC} (NaCGEC-dev)  &  449  &  1.1  &  56.1  \\

    \textbf{EXCGEC} (NLPCC-test)  &  1,611  &  1.7  &  28.9  \\

    \textbf{EXCGEC} (HSK)  &  1,824  &  2.1  &  32.0  \\

    \midrule

    \textbf{EXCGEC-train}  &  5,966  &  2.0  &  38.7  \\

    \textbf{EXCGEC-dev}    &  750    &  2.0  &  38.9  \\

    \textbf{EXCGEC-test}   &  1,500  &  2.0  &  39.2  \\

    \hdashline

    \textbf{EXCGEC} (all)  &  8,216  &  2.0  &  38.8  \\
    \bottomrule
\end{tabular}
\caption{Dataset statistics of the~\Benchmark~benchmark.}
\label{tab:statistics}
\end{table}

%% file: chapters/experiments.tex
\section{Experiments}\label{sec:experiments}

\input{tables/main_results}

\input{tables/main_results_gt}

\subsection{Experimental Settings}
\paragraph{Backbones.}
We benchmark mainstream LLMs including Qwen-1.5~\cite{bai2023qwen}, Llama-3~\cite{touvron2023llama}, and DeepSeek~\cite{bi2024deepseek}. For these LLMs, we experiment with their base and chat (or instruct) versions to investigate whether further alignment training benefits the task. All experimental results are averaged over three runs with different random seeds on EXCGEC-test in Table~\ref{tab:statistics}. More training details are reported in the supplementary materials.

\paragraph{Evaluation.}
We obtain the metric results using public toolkits including \textit{ROUGE}~\cite{lin-2004-rouge}, \textit{NLTK}~\cite{bird-loper-2004-nltk}, and \textit{scikit-learn}~\cite{scikit-learn}. Particularly, we observe many hypothesis edits are not covered by the corresponding reference edits, making it impossible to subsequently evaluate the explanations for these edits. To address this, we introduce two extra indicators, namely \textit{Hit} and \textit{Miss} rates. A hypothesis edit overlapping with a reference edit is designated as a hit edit, while a reference edit without any match with hypothesis edits is deemed a miss edit. The hit rate is defined as the ratio of hit edits to all hypothesis edits, and the miss rate as the ratio of miss edits to all reference edits. Only the hit edits are used to calculate the evaluation outcomes for explanations.


\subsection{Results of Multi-task Models}
\label{subsec:main_results}
Table~\ref{tab:main} presents the main results of multi-task models.

\paragraph{Post-explaining models outperform pre-explaining models.} Concerning the correction aspect, all post-explaining models consistently obtain higher F$_{0.5}$ scores than pre-explaining models, regardless of the applied backbones. A similar pattern is observed in the explanation part, where all the pre-explaining models invariably underperform their post-explaining counterparts. This suggests complexity for LLMs to directly explain grammatical errors without auxiliary information like target sentences or extracted edits. And once pre-explaining models generate flawed explanations, the ensuing distraction impedes their ability to accurately correct the source text.

\paragraph{Chat models are superior to base models.} For post-explaining models, we observe all chat or instruct models gain slightly higher F$_{0.5}$ correction scores, and they also marginally outperform their base version counterparts in the explanation task. It indicates that additional alignment training~\cite{wang2023aligning} can benefit the~\Task~task.

\input{tables/ablation_single_task}

\subsection{Ground Truth Results}
\label{subsec:gt_results}
To examine the isolated performance of multi-task models, we introduce partial ground truth information in advance during the formal inference stage. This is achieved by pre-inserting ground truth corrections or explanations into the decoding phase prior to formal inference. Specifically, we utilize ground truth target texts for post-explaining and evaluate the performance of the explanation task. Conversely, we provide ground truth explanations for pre-explaining and assess the performance of the correction task. This approach enables a detailed analysis of each task's performance under oracle conditions. The results, as depicted in Table~\ref{tab:main_gt}, reveal that the incorporation of ground truth information significantly enhances performance. Notably, post-explanatory models equipped with ground truth corrections exhibit a marked improvement in explanatory performance across all LLMs. This observation extends to post-explanatory models with ground truth explanations, suggesting that previously generated low-quality content adversely affects subsequent generative processes.




\subsection{Comparison with Pipeline}
We compare multi-task models and a GEC-GEE pipeline with COTE in Table~\ref{tab:ablation_single-task}. It indicates that the pipeline solution can improve both the correction and the explanation performance compared to multi-task models, highlighting the challenges of learning multi-task models for EXCGEC. However, adopting the pipeline solution requires heavy deployment and training costs. We speculate that LLMs with only 7B parameters cannot establish intimate interaction of correction and explanation tasks. 

\section{Analysis}\label{sec:analysis}

\subsection{Ablation Results}
We conduct ablation studies on Qwen1.5-7B-chat to provide in-depth insights into post-explaining models. We also study the effect of model sizes and provide a case study for different LLMs in the supplementary materials.


\input{tables/ablation_cote}

\paragraph{Effect of COTE.} We introduce COTE that provides gold alignment for post-explaining models, thus unburdening LLMs during the inference stage. The impact of COTE is quantitatively examined in this section. We provide the post-explaining model with ground truth target texts, which allows us to focus on the explanation performance. The results presented in Table~\ref{tab:ablation_cote} reveal a huge performance drop if we do not leverage COTE, especially the hit rate and the miss rate. This demonstrates the effectiveness of COTE.


\subsection{Human Evaluation for Error Descriptions}\label{subsec:human_exp}
\input{tables/human_evaluation}
We adopt traditional metrics for assessing the quality of generated error descriptions mainly for their reproductivity and efficiency~\cite{clinciu-etal-2021-study}. However, their reliability requires further validation. Therefore, this section attempts to demonstrate the suitability of these metrics through their corrections with human judgments. We assign two human annotators to score the error descriptions generated by all 6 post-explaining models, with the scoring scale from 0 to 100. For each sample, the annotators are instructed to concurrently evaluate all the error descriptions, referencing a gold explanation generated by GPT-4 to guarantee a rigorous and reliable assessment. Additional details are delineated in the supplementary materials.

We report Pearson and Spearson correlations between the metric results and the human judgments in Table~\ref{tab:human_evaluation}. We observe the inter-annotator correlations are close to 1, meaning it is relatively easy to determine the quality of error descriptions for human annotators. Most metrics achieve moderate or high correlations with human judgments, which means that it is relatively reasonable to use simple n-grams-based metrics to evaluate the quality of error descriptions efficiently. Among various metrics, ROUGE-1 achieves the highest correlations, followed by METEOR. All the introduced metrics show moderate or high correlations, indicating that it is advisable to employ them as proxies for human evaluation. We provide detailed annotation guidance and rating rules in the supplementary materials.


%% file: tables/main_results.tex
\begin{table*}[htb!]
\renewcommand{\arraystretch}{1.0}
\renewcommand{\tabcolsep}{3pt}
\centering
\scriptsize
\begin{tabular}{clcccccccccc}
\toprule

& \multicolumn{1}{c}{\multirow{2}{*}{\textbf{Model}}}
& \multicolumn{2}{c}{\textbf{Correction}$\uparrow$}
& \multicolumn{8}{c}{\textbf{Explanation}}  \\

\cmidrule(lr){3-4} \cmidrule(l){5-12} 

&
& \multicolumn{1}{c}{\textbf{CLEME (P / R / F$_{0.5}$)}}
& \multicolumn{1}{c}{\textbf{ChERRANT (P / R / F$_{0.5}$)}}
& \multicolumn{1}{c}{\textbf{Hit}$\uparrow$}
& \multicolumn{1}{c}{\textbf{Miss}$\downarrow$}
& \multicolumn{1}{c}{\textbf{Acc}$\uparrow$}
& \multicolumn{1}{c}{\textbf{F1}$\uparrow$}
& \multicolumn{1}{c}{\textbf{MAE}$\downarrow$}
& \multicolumn{1}{c}{\textbf{BLEU}$\uparrow$}
& \multicolumn{1}{c}{\textbf{METEOR}$\uparrow$}

& \multicolumn{1}{c}{\textbf{ROUGE- (1 / 2 / L)}$\uparrow$}
\\

\midrule

\multirow{6}{*}{\textbf{Post}}

& \textbf{Qwen1.5-7B-base}  
& 26.00 / 26.54 / \textbf{26.10} & 33.87 / \textbf{20.16} / 29.81 & 67.29 & 56.81 &  60.99 & \textbf{29.82} &  0.80 & 15.22 & \textbf{39.05} & 49.74 / 23.28 / 34.32     \\

& \textbf{Qwen1.5-7B-chat} 
& \textbf{28.31} / 21.21 / \textbf{26.54} & \textbf{36.74} / 17.26 / \textbf{29.98} & 68.94 & 64.83 &  \textbf{61.98} & 29.62 &  \textbf{0.75} & \textbf{15.49} & 38.88 & \textbf{50.32} / \textbf{24.25} / \textbf{35.24}     \\

\cdashline{2-12}[1pt/1pt]

& \textbf{Llama3-8B-base} 
& 20.92 / 23.60 / 21.40 & 28.81 / 17.78 / 25.63 & 61.54 & 58.38 &  58.39 & 25.12 &  0.91 & 14.54 & 37.84 & 49.53 / 23.19 / 34.58     \\

& \textbf{Llama3-8B-instruct} 
& 21.33 / 26.05 / 22.14 & 29.00 / 19.40 / 26.39 & 61.40 & \textbf{55.71} &  59.16 & 25.63 &  0.88 & 14.70 & 36.89 & 49.41 / 23.54 / 34.87     \\

\cdashline{2-12}[1pt/1pt]

& \textbf{DeepSeek-7B-base}
& 26.21 / 7.00 / 16.92 & 36.00 / 7.04 / 19.75 & \textbf{69.92} & 85.39 &  60.64 & 26.47 &  0.79 & 15.07 & 38.05 & 50.19 / 24.10 / 34.90     \\

& \textbf{DeepSeek-7B-chat} 
& 25.46 / 18.51 / 23.68 & 34.02 / 15.75 / 27.62 & 67.52 & 66.64 & 58.11 &  24.45 & 0.84 & 13.94 & 36.97 & 48.66 / 22.70 / 34.23     \\

\hline\hline

\multirow{3}{*}{\textbf{Pre}}

& \textbf{Qwen1.5-7B-chat}  
& \textbf{13.76} / \textbf{13.42} / \textbf{13.69} & \textbf{19.27} / \textbf{9.93} / \textbf{16.22} & \textbf{29.49} & 80.24 &  23.35 & 8.22 &  \textbf{1.17} & \textbf{7.75} & \textbf{27.67} & \textbf{40.47} / \textbf{15.00} / \textbf{28.20}     \\

& \textbf{Llama3-8B-instruct}  
& 7.12 / 11.17 / 7.68 & 10.86 / 8.57 / 10.31 & 23.88 & \textbf{73.06} &  \textbf{24.31} & \textbf{8.78} &  1.21 & 5.78 & 23.07 & 37.57 / 13.47 / 27.19     \\

& \textbf{DeepSeek-7B-chat}  
& 9.93 / 8.26 / 9.55 & 14.28 / 7.07 / 11.86 & 24.72 & 78.67 &  19.12 & 5.84 &  1.29 & 5.91 & 23.95 & 37.59 / 13.11 / 26.78     \\

\bottomrule
\end{tabular}
\caption{\label{tab:main}
Main results of multi-task learning models. Results of post-explaining models are listed in the \textit{top} block, while those of pre-explaining models are in the \textit{bottom} block.}
\end{table*}

%% file: tables/main_results_gt.tex
\begin{table*}[htb!]
\renewcommand{\arraystretch}{1.0}
\renewcommand{\tabcolsep}{3pt}
\centering
\scriptsize
\begin{tabular}{lcccccccccc}
\toprule

\multicolumn{1}{c}{\multirow{2}{*}{\textbf{Model}}}
& \multicolumn{2}{c}{\textbf{Correction}$\uparrow$}
& \multicolumn{8}{c}{\textbf{Explanation}}  \\

\cmidrule(lr){2-3} \cmidrule(l){4-11} 

& \multicolumn{1}{c}{\textbf{CLEME (P / R / F$_{0.5}$)}}
& \multicolumn{1}{c}{\textbf{ChERRANT (P / R / F$_{0.5}$)}}
& \multicolumn{1}{c}{\textbf{Hit}$\uparrow$}
& \multicolumn{1}{c}{\textbf{Miss}$\downarrow$}
& \multicolumn{1}{c}{\textbf{Acc}$\uparrow$}
& \multicolumn{1}{c}{\textbf{F1}$\uparrow$}
& \multicolumn{1}{c}{\textbf{MAE}$\downarrow$}
& \multicolumn{1}{c}{\textbf{BLEU}$\uparrow$}
& \multicolumn{1}{c}{\textbf{METEOR}$\uparrow$}

& \multicolumn{1}{c}{\textbf{ROUGE- (1 / 2 / L)}}
\\

\midrule

\textbf{Qwen1.5-7B-chat}  
& 62.59 / 87.35 / 66.35 & 67.58 / 69.53 / 67.96 & \textbf{99.93}  & 0.43 &  81.53 & 39.56 &  0.73 & 17.88 & 41.40 & 51.73 / 28.81 / 36.51     \\

\textbf{Llama3-8B-instruct} 
& \textbf{69.10} / \textbf{90.90} / \textbf{72.58} & \textbf{73.75} / \textbf{74.37} / \textbf{73.87} & 99.63  & 1.67 &  \textbf{85.99} & 41.84 &  0.78 & 20.73 & 42.98 & \textbf{54.60} / \textbf{29.64} / \textbf{40.04}    \\

\textbf{DeepSeek-7B-chat} 
& 41.12 / 79.02 / 45.48 & 48.35 / 53.20 / 49.25 & \textbf{99.93}  & \textbf{0.40} &  81.17 & 35.93 &  0.74 & 19.57 & 42.32 & 53.12 / 28.03 / 38.59    \\

\bottomrule
\end{tabular}
\caption{\label{tab:main_gt}
Ground truth results of multi-task learning models. We report the explanation performance (\textbf{right} block) of \textit{post-explaining} models conditioned on source texts and ground truth target texts. Contrarily, we report the correction performance (\textbf{left} block) of \textit{pre-explaining} models conditioned on source sentences and ground truth explanations.}
\end{table*}

%% file: tables/ablation_single_task.tex
\begin{table*}[!tb]
\renewcommand{\arraystretch}{1.0}
\renewcommand{\tabcolsep}{3pt}
\centering
\scriptsize
\begin{tabular}{lcccccccccc}
\toprule

\multicolumn{1}{c}{\multirow{2}{*}{\textbf{Model}}}
& \multicolumn{2}{c}{\textbf{Correction}$\uparrow$}
& \multicolumn{8}{c}{\textbf{Explanation}}  \\

\cmidrule(lr){2-3} \cmidrule(l){4-11} 

& \multicolumn{1}{c}{\textbf{CLEME (P / R / F$_{0.5}$)}}
& \multicolumn{1}{c}{\textbf{ChERRANT (P / R / F$_{0.5}$)}}
& \multicolumn{1}{c}{\textbf{Hit}$\uparrow$}
& \multicolumn{1}{c}{\textbf{Miss}$\downarrow$}
& \multicolumn{1}{c}{\textbf{Acc}$\uparrow$}
& \multicolumn{1}{c}{\textbf{F1}$\uparrow$}
& \multicolumn{1}{c}{\textbf{MAE}$\downarrow$}
& \multicolumn{1}{c}{\textbf{BLEU}$\uparrow$}
& \multicolumn{1}{c}{\textbf{METEOR}$\uparrow$}

& \multicolumn{1}{c}{\textbf{ROUGE- (1 / 2 / L)}}  \\

\midrule

\textbf{Post-explaining}
& 28.31 / 21.21 / 26.54 & 36.74 / 17.26 / 29.98 & 68.94 & 64.83 &  61.98 & 29.62 &  \textbf{0.75} & 15.49 & 38.88 & 50.32 / 24.25 / 35.24     \\

\textbf{Pre-explaining}
& 13.76 / 13.42 / 13.69 & 19.27 / 9.93 / 16.22 & 29.49 & 80.24 &  23.35 & 8.22 &  1.17 & 7.75 & 27.67 & 40.47 / 15.00 / 28.20     \\



\textbf{GEC-GEE Pipeline} 
& \textbf{32.45} / \textbf{23.93} / \textbf{30.29} & \textbf{40.50} / \textbf{19.58} / \textbf{33.37} & \textbf{72.00} & \textbf{63.10} & \textbf{65.76} & \textbf{32.77} &  \textbf{0.70} & \textbf{16.41} & \textbf{40.04} & \textbf{51.07} / \textbf{24.92} / \textbf{35.89}     \\

\bottomrule
\end{tabular}
\caption{\label{tab:ablation_single-task}
Comparison of the multi-task solutions and the GEC-GEE pipeline solution based on Qwen1.5-7B-chat.}
\end{table*}

%% file: tables/ablation_cote.tex
\begin{table}[!tb]
\renewcommand{\arraystretch}{1.0}
\renewcommand{\tabcolsep}{3pt}
\centering
\scriptsize
\begin{tabular}{lcccccccc}
\toprule

& \multicolumn{1}{c}{\textbf{Hit}$\uparrow$}
& \multicolumn{1}{c}{\textbf{Miss}$\downarrow$}
& \multicolumn{1}{c}{\textbf{Acc}$\uparrow$}
& \multicolumn{1}{c}{\textbf{F1}$\uparrow$}
& \multicolumn{1}{c}{\textbf{MAE}$\downarrow$}
& \multicolumn{1}{c}{\textbf{ROUGE- (1/2/L)}$\uparrow$}
\\

\midrule

\textbf{w COTE}  
& \textbf{99.93} & \textbf{0.43} & \textbf{81.53} & \textbf{39.56} & \textbf{0.74} & \textbf{51.73} / \textbf{25.81} / \textbf{36.51}     \\

\textbf{w/o COTE} 
& 49.64 & 54.01 & 42.51 & 17.77 & 0.93 & 46.35 / 19.34 / 31.28     \\



\bottomrule
\end{tabular}
\caption{\label{tab:ablation_cote}
Ablation results of COTE from the same model.}
\end{table}

%% file: tables/human_evaluation.tex
\begin{table}[tb!]
\renewcommand{\arraystretch}{0.9}
\renewcommand{\tabcolsep}{5pt}
\centering
\scriptsize
\begin{tabular}{lcc}
\toprule

&  \textbf{Pearson}  &  \textbf{Spearson}  \\

\midrule

\textbf{Human} v.s. \textbf{BLEU}       &  0.9222  &  0.6571  \\
\textbf{Human} v.s. \textbf{METEOR}     &  0.9280  &  0.7714  \\
\textbf{Human} v.s. \textbf{ROUGE-1}    &  0.9464  &  0.8286  \\
\textbf{Human} v.s. \textbf{ROUGE-2}    &  0.9175  &  0.4857  \\
\textbf{Human} v.s. \textbf{ROUGE-L}    &  0.9352  &  0.6571  \\

\hline

\textbf{A$_1$} v.s. \textbf{A$_2$}  &  0.9874  &  0.9429  \\

\bottomrule
\end{tabular}
\caption{\label{tab:human_evaluation}
Correlations between human judgements (A$_1$ and A$_2$) and metrics results for error descriptions.}
\end{table}

%% file: chapters/conclusion.tex
\section{Conclusion}
We propose and formulate the task of~\Task, establishing the interaction of correction and explanation tasks. To develop the task, we propose the EXCGEC benchmark, based on which we build baseline models. Extensive experiments and analyses reveal several challenges of the task. We hope this paper can serve as a starting point for future exploration.

\section*{Acknowledgements}
This research is supported by National Natural Science Foundation of China (Grant No.62276154), Research Center for Computer Network (Shenzhen) Ministry of Education, the Natural Science Foundation of Guangdong Province (Grant No.2023A1515012914 and 440300241033100801770), Basic Research Fund of Shenzhen City (Grant No.JCYJ20210324120012033 and GJHZ20240218113603006), the Major Key Project of PCL for Experiments and Applications (PCL2021A06).

%% file: chapters/appendix.tex
\section{The Details of~\Benchmark~Benchmark}

\subsection{Description of Grammatical Error Types}
\label{app:error_types}

In the taxonomy of Chinese grammatical errors, we classify errors based on their level of granularity into five major types: punctuation-level, spelling-level, word-level, sentence-level, and other special errors. This section provides a detailed description of each grammatical error type.

\paragraph{Punctuation-level Error.} These errors primarily involve redundancy, omission, or misuse of punctuation marks.

\begin{itemize}

    \item 标点冗余~(\textbf{Punctuation Redundancy}). Punctuation redundancy refers to the unnecessary insertion of punctuation marks. To explain such errors, we first specify the role of the involved punctuation marks and then elaborate on why their presence is redundant in the given context.

    \item 标点丢失~(\textbf{Punctuation Missing}). Punctuation missing occurs when necessary punctuation marks are omitted, typically in the middle or at the end of a sentence. To address such errors, we identify the missing punctuation and explain its intended function within the context.
    
    \item 标点误用~(\textbf{Punctuation Misuse}). Punctuation misuse is a frequent error in everyday Chinese writing. For such cases, we briefly describe the functions of the misused and correct punctuation marks and provide a rationale for the correction.
\end{itemize}

\paragraph{Spelling-level Error.} Spelling-level errors arise when individuals due to carelessness or lack of knowledge, write incorrect characters or words. These errors are prevalent, making Chinese Spelling Check (CSC)~\cite{huang-etal-2023-frustratingly} a standard NLP task that has garnered significant attention from researchers. Inspired by these studies, we further categorize spelling-level errors into four subtypes:

\begin{itemize}
    \item 字音混淆错误~(\textbf{Phonetic Confusion Error}). Phonetic confusion errors occur when characters with similar or identical pinyin are misused. Since most Chinese Internet users utilize the pinyin input method, this type of error is particularly common in online texts.
    
    \item 字形混淆错误~(\textbf{Glyph Confusion Error}). Glyph confusion errors arise when users of glyph-based input methods, such as Wubi, confuse characters with similar fonts or strokes, leading to spelling mistakes.
    
    \item 词内部字符异位错误~(\textbf{Internal Character Misplacement Error}). Internal character misplacement errors involve the incorrect ordering of characters within a multi-character word. These errors are rare among native speakers but may occur in texts written by second-language (L2) learners. For instance, the erroneous word ``共公'' should be corrected to ``公共''.
    
    \item 命名实体拼写错误~(\textbf{Named Entity Misspelling}). There are numerous named entity words in Chinese, such as person names, organization names, place names, and all other entities identified by terminologies. These words are also very prone to spelling errors.
\end{itemize}

\paragraph{Word-level Error.} Word-level errors involve the improper use of individual words or idioms within a sentence, even when the sentence's syntactic structure is correct. These errors are among the most common in Chinese texts and can be further divided into three subtypes:

\begin{itemize}
    \item 词语冗余~(\textbf{Word Redundancy}). Word redundancy occurs when words with identical or similar meanings appear together in a sentence, resulting in semantic repetition and redundancy. Such words are often adjacent, so it is essential to examine whether their meanings overlap, potentially causing redundancy.
    
    \item 词语丢失~(\textbf{Word Missing}). In modern Chinese, a sentence typically comprises six major components: subject, predicate, object, attributive, adverbial, and complement. While not all components are required in every sentence, the sentence must include the necessary elements to convey a complete meaning. The omission of essential components results in word-missing errors.
    
    \item 词语误用~(\textbf{Word Misuse}). Word misuse refers to the inappropriate use of words in a sentence. This error often stems from the author's insufficient understanding of a word's meaning or grammatical role.
\end{itemize}

\paragraph{Sentence-level Error.} Sentence-level errors occur when the grammatical structure of a sentence is violated or when logical reasoning is not followed. These errors can be categorized into three subtypes:

\begin{itemize}
    \item 词序不当~(\textbf{Improper Word Order}). Proper word order is crucial for conveying precise meaning in Chinese. Incorrect word order disrupts sentence structure, leading to confusion and imbalance between sentence components, ultimately affecting meaning.
    
    \item 逻辑不通~(\textbf{Illogicality}). Illogicality occurs when a sentence adheres to grammatical norms but violates logical reasoning. This can result from issues such as improper logical sequencing, causal confusion, or the reversal of subject and object.
    
    \item 句式杂糅~(\textbf{Run-on Sentence}). A run-on sentence in Chinese typically involves the blending of two formats or sentences with similar meanings into one. Writers may begin with one format but unconsciously switch to another due to interference from content or other factors, thus resulting in a mixed structure.
\end{itemize}

\paragraph{Other Special Error.} In addition to the aforementioned types, certain errors do not neatly fit into these categories. These are classified as other special errors and can be further subdivided into three types:

\begin{itemize}
    \item 照应错误~(\textbf{Inconsistency Error}). Inconsistency errors involve incorrect referential relationships between words. Identifying and explaining such errors requires an understanding of these relationships.
    
    \item 歧义错误~(\textbf{Ambiguity Error}). Ambiguity errors arise when a word or sentence can be interpreted in multiple ways, leading to confusion.
    
    \item 语气不协调~(\textbf{Inconsistent Tone}). Inconsistent tone occurs when the tone or style of preceding and subsequent sentences is mismatched.
\end{itemize}

Additionally, we classify errors that do not align with any of the aforementioned categories as ``\textit{Other}''. These errors often require substantial revisions and may involve alterations to the original semantics.

\input{figures/error_type}

\subsection{Description of Grammatical Error severity}
\label{app:severity}
In our work, the severity of grammatical errors is categorized into five levels, ranging from 1 to 5 points. Each level is defined as follows:

\begin{itemize}
    \item \textbf{1 point~(\textit{Trivial Error})}: These are minor issues, such as typing errors or slight word misuse, with minimal impact on the overall meaning. Example: 他十分擅长数学英语'' should be corrected to 他十分擅长数学和英语''.

    \item \textbf{2 points~(\textit{Minor Error})}: These errors may slightly obscure the intended expression but do not hinder overall comprehension. Example: ``我喜欢狗和录像游戏'' should be corrected to ``我喜欢狗和电子游戏''.

    \item \textbf{3 points~(\textit{Moderate Error})}: These errors can cause parts of the sentence to become incoherent, requiring the reader to reread or pause to understand the intended meaning. Example: ``我走家去了'' should be corrected to ``我走去家了''.
    
    \item \textbf{4 points~(\textit{Serious Error})}: These errors not only disrupt comprehension but may also significantly alter the intended meaning of the sentence. Example: ``我想借用你的手机扮演职业摄影师'' should be corrected to ``我想借用你的手机拍摄一些专业的照片''.
    
    \item \textbf{5 points~(\textit{Extremely Serious Error})}: These errors render the sentence nearly or completely incomprehensible. Example: ``他举妈妈，我去购物车'' should be corrected to ``他举着妈妈的购物车，我就去了''.
\end{itemize}

\input{tables/hyper_parameters}

\input{tables/ablation_model_sizes}

\input{tables/ablation_decoding}

\subsection{Examples of Error Types}
\label{app:examples}
Figure~\ref{fig:error_type_details_1},~\ref{fig:error_type_details_2}, and~\ref{fig:error_type_details_3} list the examples of all the error types involved in this paper.

\subsection{Prompt of Generating Explanations}
\label{app:prompt}
The prompt we use to generate explanations is shown in Figure~\ref{fig:prompt}. with its English version in Figure~\ref{fig:prompt_eng}.

\subsection{Detailed Description of LLM Failure Modes}
\label{app:failure_modes}

We categorize the failure modes of LLMs into seven principal types: incorrect type, incorrect severity, incorrect format, incorrect template, non-fluency, unreasonability, and unfaithfulness. To analyze these failure modes, an expert annotator reviewed 100 randomly sampled invalid explanations, with each explanation potentially falling into multiple categories. The annotation results presented in Figure~\ref{fig:error_type} reveal that GPT-4 frequently misclassifies grammatical errors and provides unfaithful error descriptions. However, GPT-4 demonstrates notable strengths in generating well-structured, fluent, and reasonable explanations, highlighting the potential of utilizing LLM annotations for this task. Below, we provide detailed definitions for each of the seven failure modes:

\begin{itemize}
    \item \textbf{Incorrect Type}: The identified error type does not match the actual grammatical issue.
    
    \item \textbf{Incorrect Format}: The evidence and correction content are not properly highlighted using designated markers including 【~】 for evidence or \{~\} for correction.

    \item \textbf{Incorrect Template}: The error description fails to adhere to the structured syllogism format based on deductive reasoning.

    \item \textbf{Non-Fluency}: The error description is poorly written, non-fluent, or difficult to read.

    \item \textbf{Unreasonability}: The error description contains clear linguistic inaccuracies, making it unacceptable or implausible for human understanding.
    
    \item \textbf{Unfaithfulness}: The error description does not accurately address or correspond to the specific edit.
\end{itemize}

\section{Experimental Details and Extra Results}

\subsection{Implementation Details.}
\label{app:implementation}
Each model undergoes training for a total of five epochs, after which the optimal model is identified through validation on EXCGEC-dev. The performance of this best-performing model is subsequently evaluated on EXCGEC-test. Comprehensive details of the training hyperparameters for all models in our study are provided in Table~\ref{tab:hyper}.

\subsection{Effect of Model Sizes}
\label{app:model_sizes}
Table~\ref{tab:ablation_model_sizes} indicates the varying performance across model sizes ranging from 1.8B to 7B. We observe consistent performance enhancement with increasing model sizes.

\subsection{Effect of token-wise decoding strategies}
By default, we employ beam search decoding for corrections and Top-p decoding for explanations. In this section, we explore the reverse setting, and the results are reported in Table~\ref{tab:ablation_decoding}. When switching from beam search to top-p for correction, we observe a huge performance drop in precision and F$_{0.5}$ and an increase in recall, which means top-p encourages LLMs to over-correct~\cite{cao2023unnatural}. On the other hand, leveraging beam search improves explanation performance, suggesting the potential benefits of a greedy decoding algorithm for the task. However, we notice that beam search also increases the miss rate. We speculate that beam search may discard some low-likelihood explanations.

\subsection{Case Study}
\label{app:case_studies}
We provide a case study of the generated explanations in the JSON format by various LLMs in Figure~\ref{fig:case_study}.

\subsection{Details of Human Rating}
\label{app:human_rating}

Specifically, we hire 2 native Chinese speakers to rate the explanations generated by 6 post-explaining models in Table~\ref{tab:main} conditioned on ground truth target texts. The rating scores range from 0 to 100, and each annotator concurrently rates 6 explanations for each sample. We randomly select 100 samples for annotation. We provide annotators with general scoring suggestions:

\begin{itemize}
    \item \textbf{100 points}: The explanations are highly fluent, incorporate relevant semantic knowledge to enhance persuasiveness (rationality), and demonstrate clear alignment with the current editor's context (loyalty). All aspects are impeccable, leaving little to no room for improvement in terms of explanation or description.

    \item \textbf{80$\sim$100 points}: The explanations are fluent, meet the criteria for fidelity, and exhibit a reasonable degree of rationality. However, minor flaws are present.
    
    \item \textbf{60$\sim$80 points}: The explanations are fluent, but there are noticeable shortcomings in either fidelity or rationality. While somewhat useful for understanding and correcting grammatical errors, the overall quality is limited.
    
    \item \textbf{30$\sim$60 points}: The explanations are fluent, but the rationality is weak, offering minimal assistance in understanding and correcting grammatical errors.
    
    \item \textbf{0$\sim$30 points}: The explanations are fluent, but fidelity is lacking, and the explanation fails to address the current editor's context. It provides no meaningful assistance for understanding or correcting grammatical errors.
    
    \item \textbf{0$\sim$30 points}: The explanations are vague and unintelligible, offering no value in understanding or correcting grammatical errors.
\end{itemize}

\input{figures/error_type_details}

\input{figures/prompt}

\input{figures/prompt_eng}

\input{figures/case_study}

%% file: figures/error_type.tex
\begin{figure}[tb!]
\centering
\includegraphics[scale=0.55]{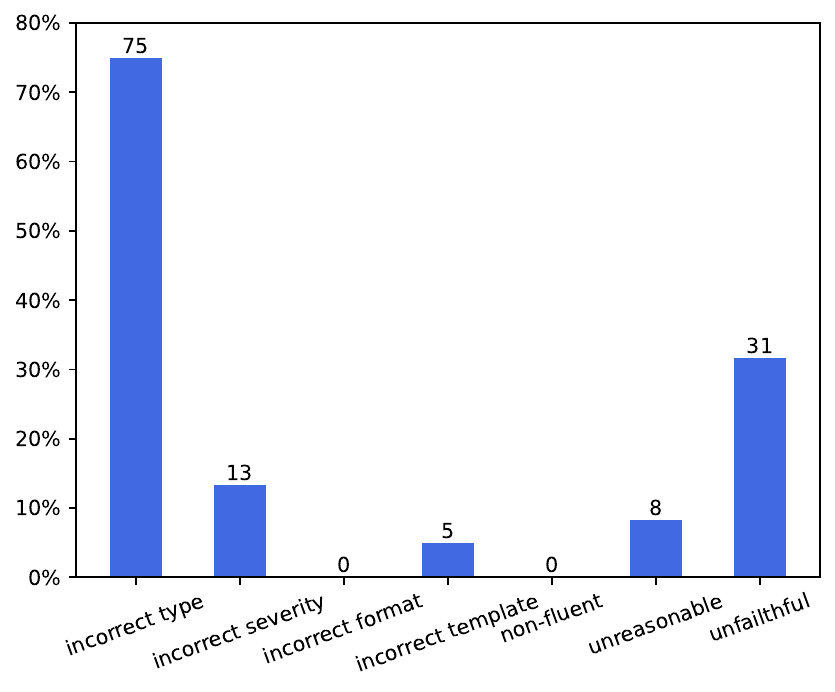}
\caption{Distribution of 7 kinds of LLM errors.}
\label{fig:error_type}
\end{figure}

%% file: tables/hyper_parameters.tex
\begin{table}[tb!]
\renewcommand{\arraystretch}{1.0}
\renewcommand{\tabcolsep}{3pt}
\centering
\scriptsize
\begin{tabular}{lc}
\toprule
\textbf{Configuration}  & \textbf{Value}    \\ 
\midrule
\multicolumn{2}{c}{\textbf{Fine-tuning}}   \\ 
\midrule

Devices   &  2 Tesla A100 GPU (80GB)        \\
Epochs    &  5                              \\
Finetuning type                   & Lora    \\
Train batch size per GPU          & 2       \\
Eval batch size per GPU           & 1       \\
Gradient accumulation steps       & 16      \\

Optimizer & 
\begin{tabular}[c]{@{}c@{}}
AdamW  \\ 
($\beta_1=0.9,\beta_2=0.98,\epsilon=1 \times 10^{-6}$) 
\end{tabular}    \\

Learning rate           &  $5 \times 10^{-5}$    \\
Learning rate schedule  &  cosine decay \\
Warmup steps            & 20            \\
Eval steps              & 200           \\
Cutoff length           & 1024          \\
Preprocessing workers number      & 16  \\
Numerical precision     & fp16          \\
Weight decay            & 0.05          \\

\midrule
\multicolumn{2}{c}{\textbf{Inference}}              \\ 
\midrule

Beam size               & 5     \\
Top-p                   & 0.8   \\
Max new tokens          & 2048  \\
Temperature             & 0.7   \\

\bottomrule

\end{tabular}
\caption{Hyper-parameters used in our experiments.}
\label{tab:hyper}
\label{tab:hp}
\end{table}

%% file: tables/ablation_model_sizes.tex
\begin{table*}[tbh!]
\renewcommand{\arraystretch}{1.0}
\renewcommand{\tabcolsep}{3pt}
\centering
\scriptsize
\begin{tabular}{lcccccccccc}
\toprule

\multicolumn{1}{c}{\multirow{2}{*}{\textbf{Model}}}
& \multicolumn{2}{c}{\textbf{Correction}$\uparrow$}
& \multicolumn{7}{c}{\textbf{Explanation}}  \\

\cmidrule(lr){2-3} \cmidrule(lr){4-11} 

& \multicolumn{1}{c}{\textbf{CLEME (P / R / F$_{0.5}$)}}
& \multicolumn{1}{c}{\textbf{ChERRANT (P / R / F$_{0.5}$)}}
& \multicolumn{1}{c}{\textbf{Hit}$\uparrow$}
& \multicolumn{1}{c}{\textbf{Miss}$\downarrow$}
& \multicolumn{1}{c}{\textbf{Acc}$\uparrow$}
& \multicolumn{1}{c}{\textbf{F1}$\uparrow$}
& \multicolumn{1}{c}{\textbf{MAE}$\downarrow$}
& \multicolumn{1}{c}{\textbf{BLEU}$\uparrow$}
& \multicolumn{1}{c}{\textbf{METEOR}$\uparrow$}

& \multicolumn{1}{c}{\textbf{ROUGE- (1 / 2 / L)}}
\\

\midrule


\textbf{Qwen1.5-1.8B-chat}  
& 21.11 / 19.28 / 20.72 & 28.91 / 15.70 / 24.74 & 59.94 & 65.14 & 55.80 & 23.27 &  0.89 & 10.19 & 34.35 & 48.66 / 22.70 / 34.23     \\

\textbf{Qwen1.5-4B-chat}  
& 22.49 / 20.84 / 22.14 & 30.57 / 16.85 / 26.29 & 62.91 & \textbf{62.70} & 57.16 & 25.31 &  0.85 & 11.61 & 35.91 & 46.83 / 19.59 / 30.86     \\

\textbf{Qwen1.5-7B-chat} 
& \textbf{28.31} / \textbf{21.21} / \textbf{26.54} & \textbf{36.74} / \textbf{17.26} / \textbf{29.98} & \textbf{68.94} & 64.83 & \textbf{61.98} & \textbf{29.62} &  \textbf{0.75} & \textbf{15.49} & \textbf{38.88} & \textbf{50.32} / \textbf{24.25} / \textbf{35.24}     \\

\bottomrule
\end{tabular}
\caption{\label{tab:ablation_model_sizes}
Comparison of post-explaining models with various model sizes.}
\end{table*}

%% file: tables/ablation_decoding.tex
\begin{table*}[tbh!]
\renewcommand{\arraystretch}{1.0}
\renewcommand{\tabcolsep}{3pt}
\centering
\scriptsize
\begin{tabular}{lcccccccccc}
\toprule

\multicolumn{1}{c}{\multirow{2}{*}{\textbf{Decoding}}}
& \multicolumn{2}{c}{\textbf{Correction}$\uparrow$}
& \multicolumn{8}{c}{\textbf{Explanation}}  \\

\cmidrule(lr){2-3} \cmidrule(l){4-11}

& \multicolumn{1}{c}{\textbf{CLEME (P / R / F$_{0.5}$)}}
& \multicolumn{1}{c}{\textbf{ChERRANT (P / R / F$_{0.5}$)}}
& \multicolumn{1}{c}{\textbf{Hit}$\uparrow$}
& \multicolumn{1}{c}{\textbf{Miss}$\downarrow$}
& \multicolumn{1}{c}{\textbf{Acc}$\uparrow$}
& \multicolumn{1}{c}{\textbf{F1}$\uparrow$}
& \multicolumn{1}{c}{\textbf{MAE}$\downarrow$}
& \multicolumn{1}{c}{\textbf{BLEU}$\uparrow$}
& \multicolumn{1}{c}{\textbf{METEOR}$\uparrow$}

& \multicolumn{1}{c}{\textbf{ROUGE- (1 / 2 / L)}}  \\

\midrule

\textbf{Beam Search}  
& \textbf{28.31} / 21.21 / \textbf{26.54} & \textbf{36.74} / 17.26 / \textbf{29.98} & 99.22 & 19.05 & \textbf{83.93} & \textbf{44.48} &  \textbf{0.71} & \textbf{22.71} & \textbf{44.28} & \textbf{55.55} / \textbf{32.26} / \textbf{42.34}     \\

\textbf{Top-p} 
& 19.45 / \textbf{27.05} / 20.61 & 24.83 / \textbf{19.14} / 23.44 & \textbf{99.93} & \textbf{0.40} & 81.53 & 39.56 &  0.74 & 17.88 & 41.40 & 51.73 / 25.81 / 36.51    \\

\bottomrule
\end{tabular}
\caption{\label{tab:ablation_decoding}
Comparison of the post-explaining model with different token-wise decoding strategies. Note that the explanation performance is conditioned on ground truth target texts to exclude unrelated interference.}
\end{table*}

%% file: figures/error_type_details.tex
\begin{figure*}[tb!]
\centering
\includegraphics[scale=0.28]{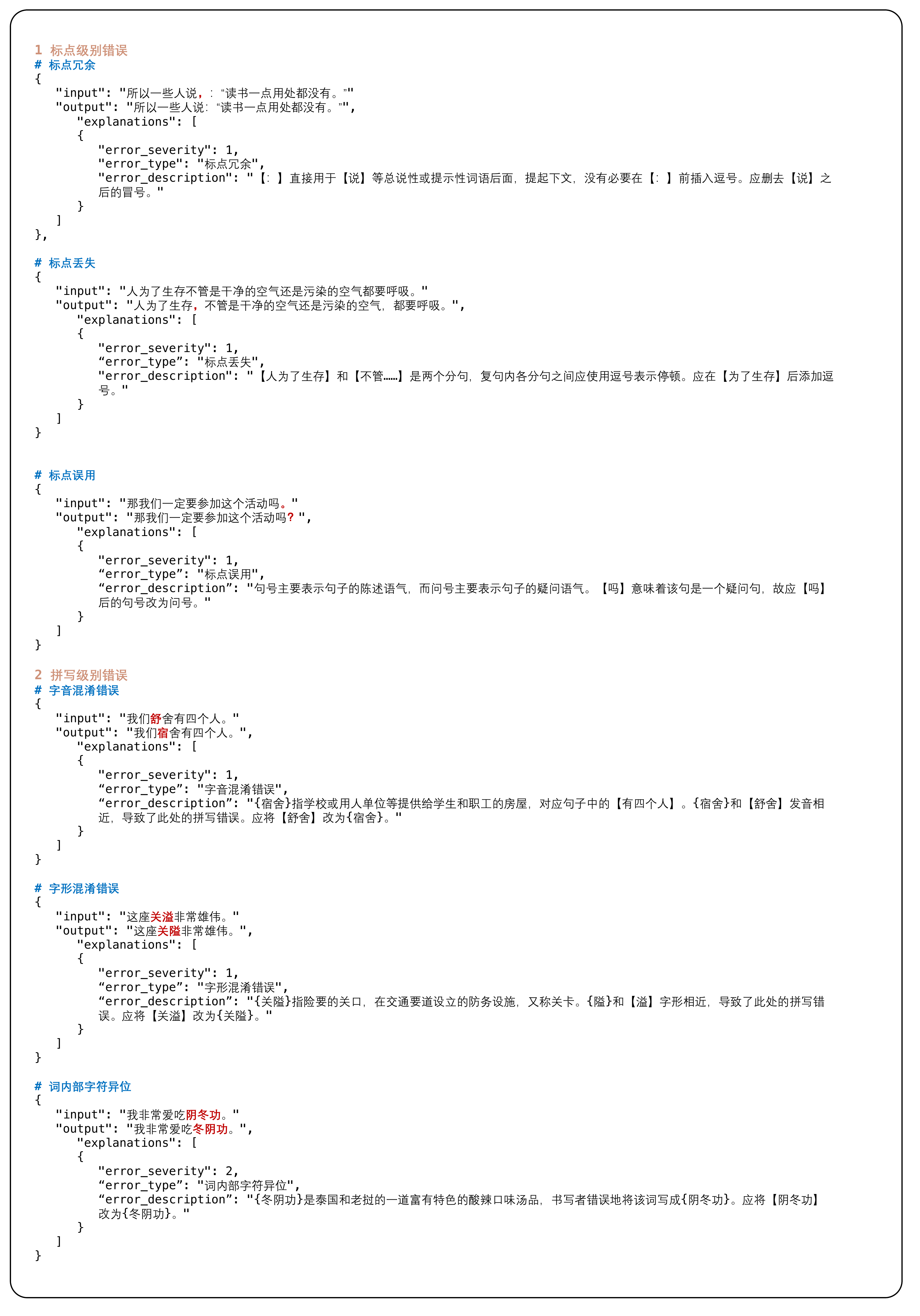}
\caption{Examples of error types.}
\label{fig:error_type_details_1}
\end{figure*}

\begin{figure*}[tb!]
\centering
\includegraphics[scale=0.28]{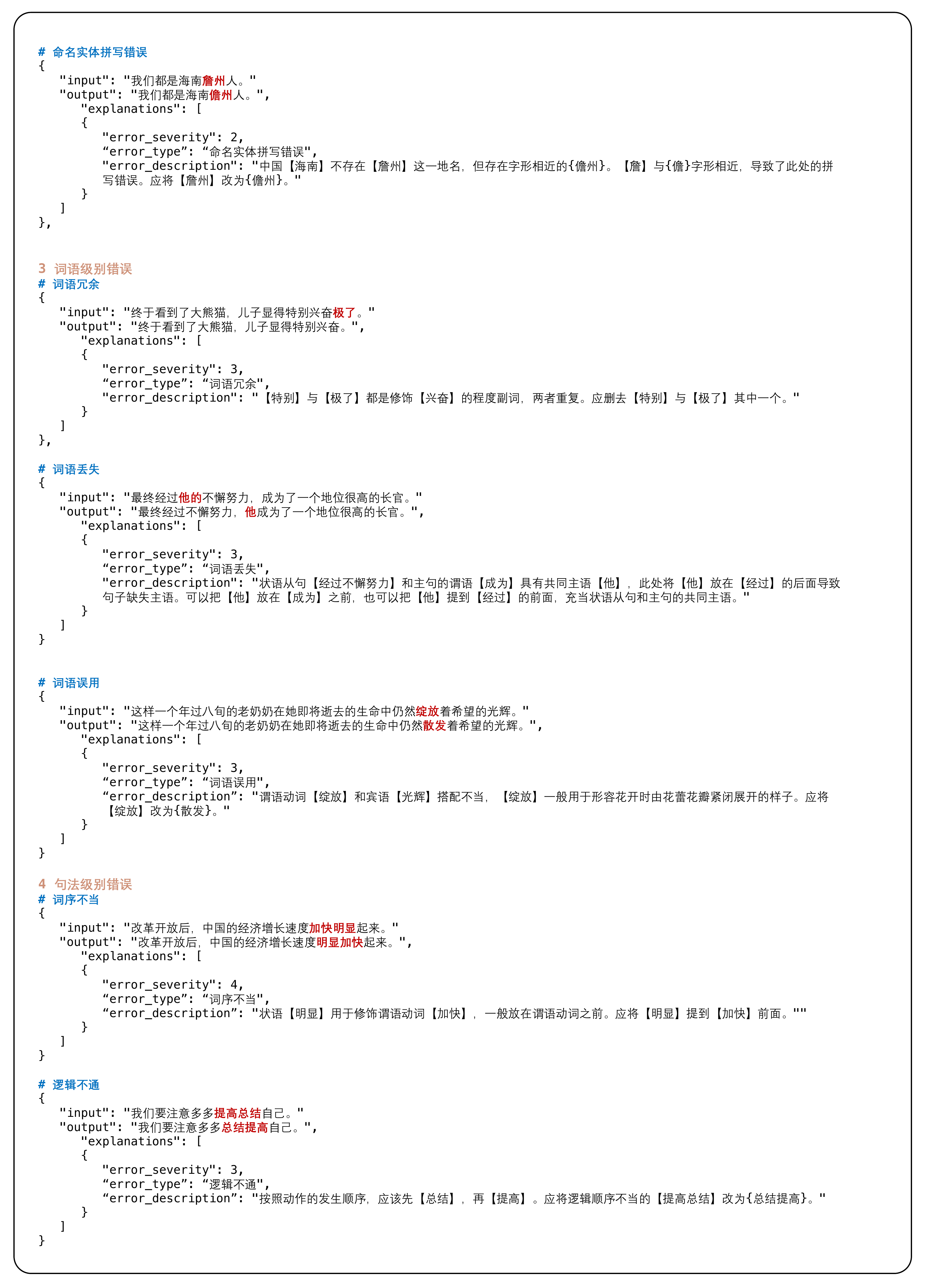}
\caption{Examples of error types.}
\label{fig:error_type_details_2}
\end{figure*}

\begin{figure*}[tb!]
\centering
\includegraphics[scale=0.28]{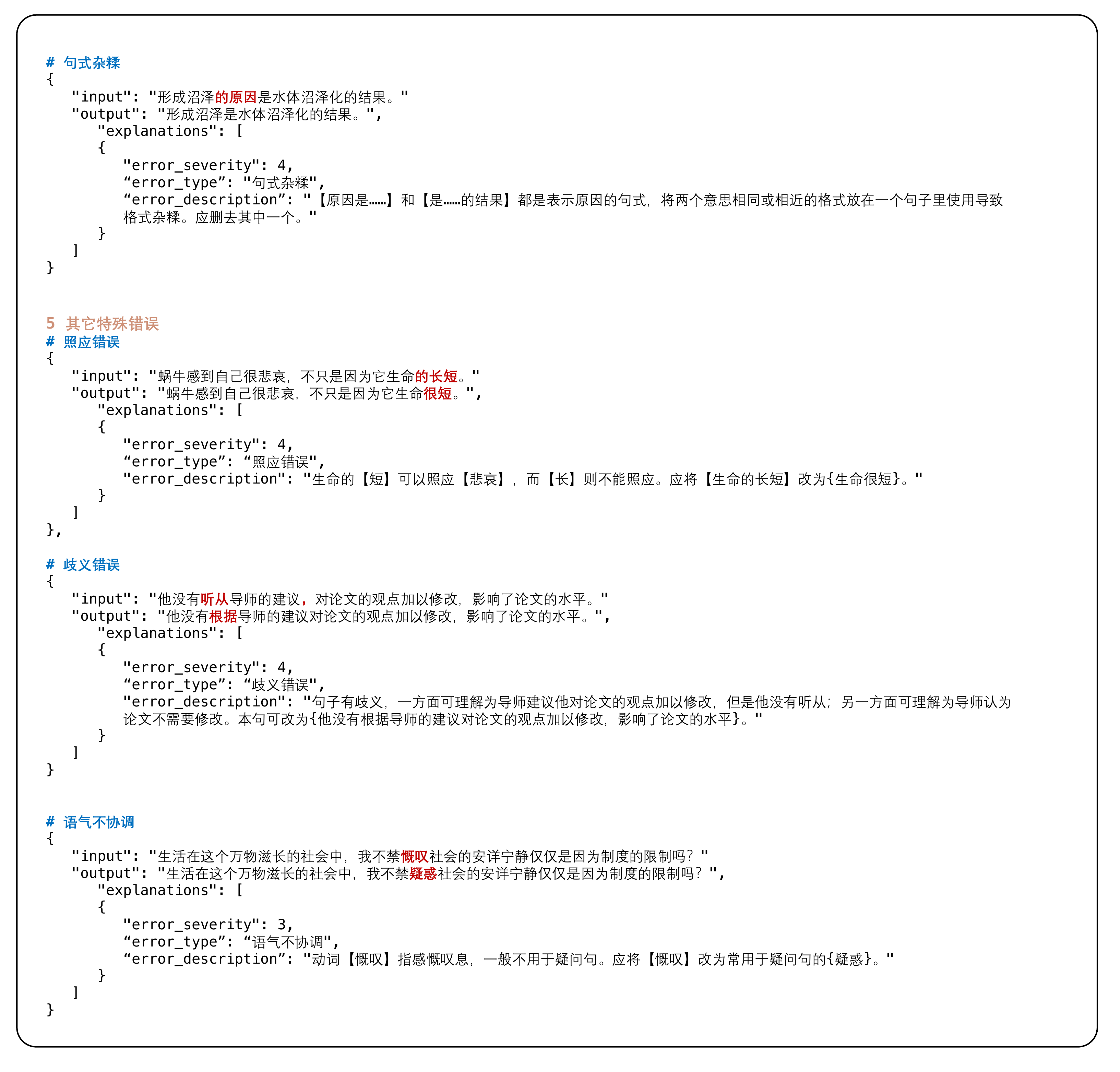}
\caption{Examples of error types.}
\label{fig:error_type_details_3}
\end{figure*}

%% file: figures/prompt.tex
\begin{figure*}[tb!]
\centering
\includegraphics[scale=0.28]{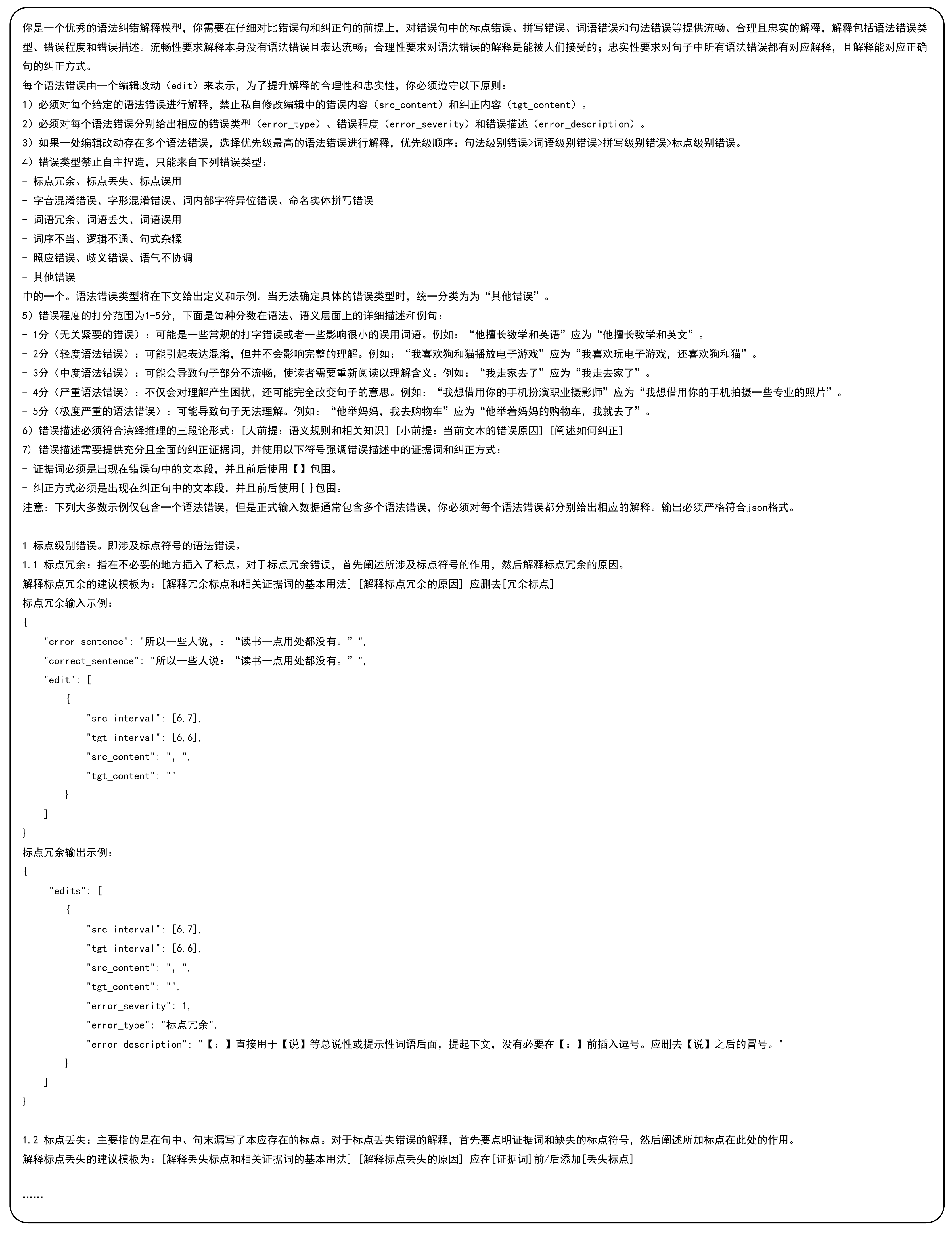}
\caption{The prompt used for explanation generation. For each error type, We provide the definition, a suggested template of error description, and a demonstration for GPT-4.}
\label{fig:prompt}
\end{figure*}

%% file: figures/prompt_eng.tex
\begin{figure*}[tb!]
\centering
\includegraphics[scale=0.265]{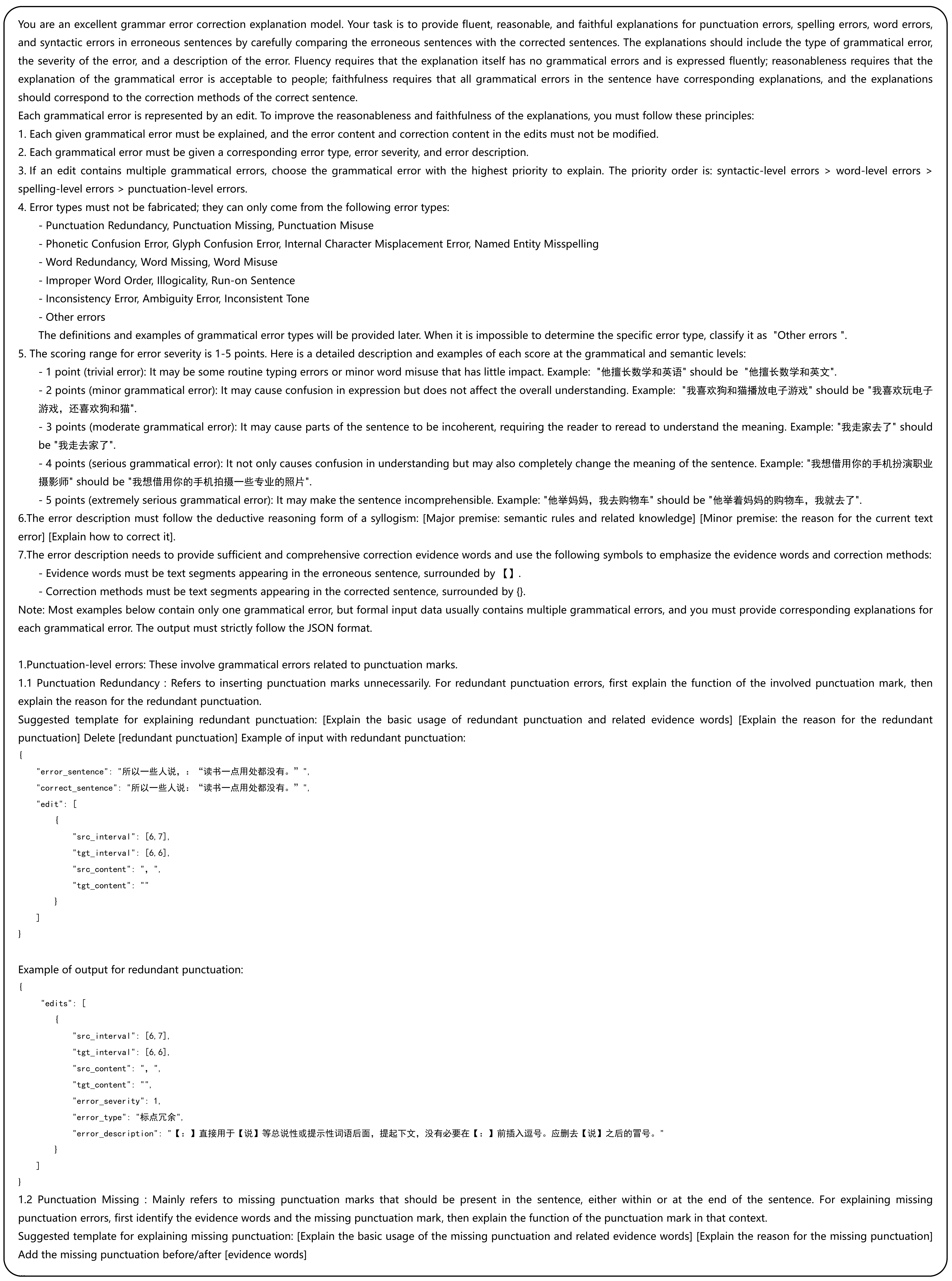}
\caption{The English prompt used for explanation generation.}
\label{fig:prompt_eng}
\end{figure*}

%% file: figures/case_study.tex
\begin{figure*}[tb!]
\centering
\vspace{0cm} 
\includegraphics[scale=0.28]{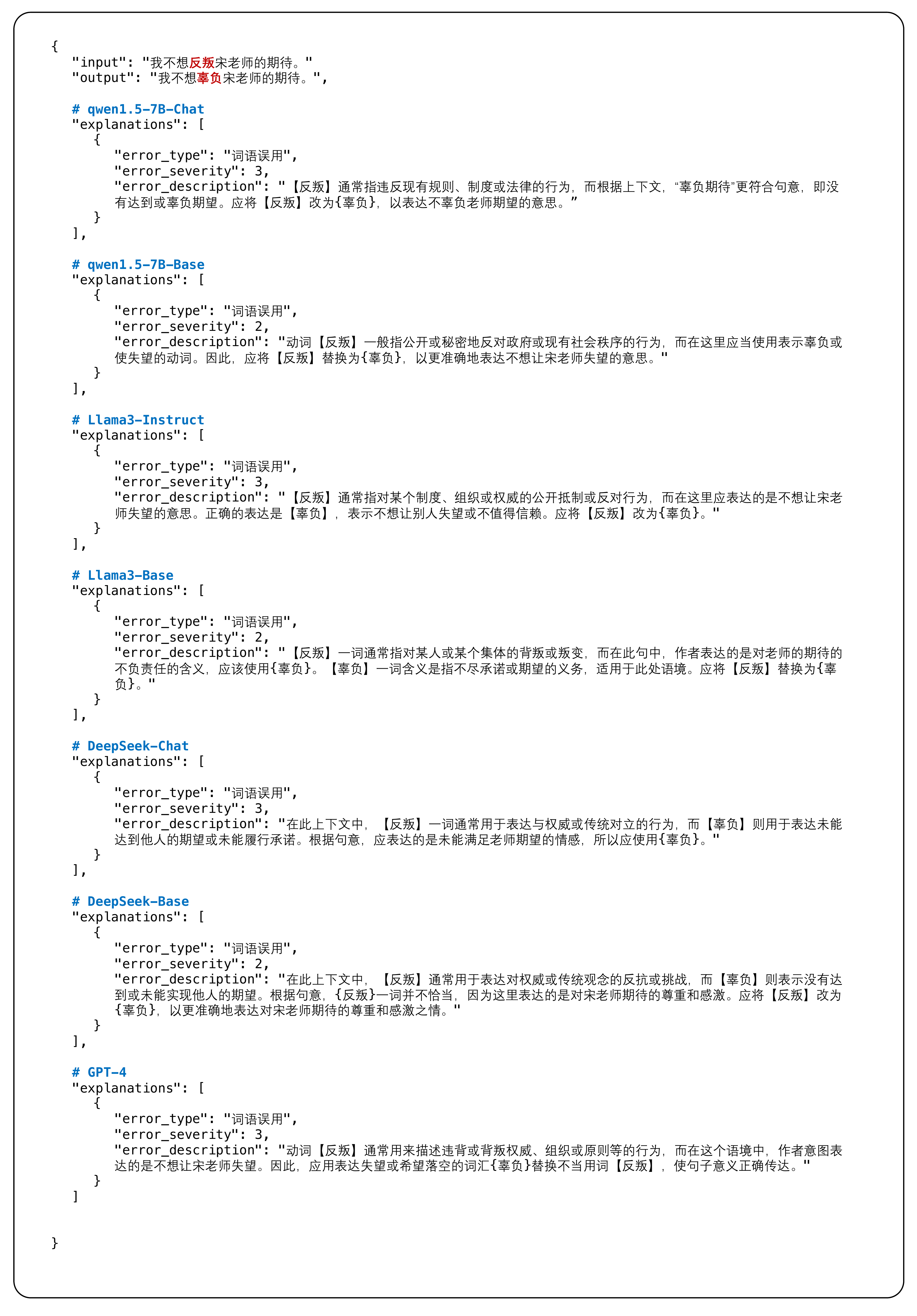}
\caption{A case study of all the LLMs involved in our experiments.}
\label{fig:case_study}
\end{figure*}